\documentclass[letter]{ieice}
\usepackage[pdftex]{graphicx,xcolor}
\usepackage[fleqn]{amsmath}
\usepackage{newtxtext}
\usepackage[varg]{newtxmath}

\usepackage{cite}
\usepackage{bm}
\usepackage{mathtools}
\usepackage{url}

\newtheorem{prop}{Proposition}
\newtheorem{theo}{Theorem}
\newtheorem{coro}{Corollary}

\field{}
\title{Batch Updating of a Posterior Tree Distribution over a Meta-Tree}
\authorlist{%
 \authorentry[y.nakahara@waseda.jp]{Yuta NAKAHARA}{}{labelA}
 \authorentry{Toshiyasu MATSUSHIMA}{}{labelB}
}
\affiliate[labelA]{The author is with the Center for Data Science, Waseda University, 1-6-1 Nishiwaseda, Shinjuku-ku, Tokyo, 162-8050, Japan.}
\affiliate[labelB]{The author is with the Department of Pure and Applied Mathematics, Waseda University, 3-4-1 Okubo, Shinjuku-ku, Tokyo, 169-8555, Japan.}

\received{2015}{1}{1}
\revised{2015}{1}{1}

\begin{document}
\maketitle
\begin{summary}
Previously, we proposed a probabilistic data generation model represented by an unobservable tree and a sequential updating method to calculate a posterior distribution over a set of trees. The set is called a meta-tree. In this paper, we propose a more efficient batch updating method.
\end{summary}

\section{Introduction}
In the field of decision trees, most previous studies (e.g., \cite{CART}) used a tree to represent a predictive function of a new data point. In contrast, we used a tree to represent a probabilistic data generation and observation model behind data\cite{suko_alg,MTRF}. Given a training data generated from that probabilistic model, we considered the Bayes optimal prediction of a new data point generated from the same model. To execute the Bayes optimal prediction, we required a posterior distribution of the tree representing the model. 

Therefore, we previously proposed an updating method\cite{suko_alg,MTRF} to calculate the posterior distribution over a set of trees called a meta-tree by applying a Bayes coding algorithm\cite{bayes_codes,CT} for context tree source in information theory. This algorithm was originally aimed to encode, transmit, and decode a sequence. In such a situation, a sequential method is preferable to reduce a delay. In \cite{suko_alg,MTRF}, this algorithm was applied to decision trees as it was. As a result, the method in \cite{suko_alg,MTRF} was also sequential, although we have little motivation to reduce such a delay in usual machine learning.

Thus, we propose a more efficient batch updating method to calculate the posterior distribution in this paper. This method includes the sequential updating method when the batch size is one. It is implemented in our open source library\cite{bayesml}.

\section{Preliminaries}

First, we introduce a notion of meta-tree\cite{MTRF}. Let $T$ denote a $M$-ary regular tree whose depth is smaller than or equal to $D_\mathrm{max}$. For any $T$, let $\mathcal{S}(T)$, $\mathcal{L}(T)$, and $\mathcal{I}(T)$ denote the set of nodes, leaf nodes, and inner nodes of $T$, respectively. Each inner node $s \in \mathcal{I}(T)$ of $T$ has a index $k_s \in \{ 1, 2, \dots , K \}$, which is called feature assignment index. Let $\bm k$ denote their tuple $(k_s)_{s \in \mathcal{I}(T)}$. For a tree $T$ (which is called representative tree) and its feature assignment indices $\bm k$, meta-tree $M_{T, \bm k}$ is defined as a set of pruned subtrees whose feature assignment indices are same as $\bm k$, i.e.,
\begin{align}
    M_{T, \bm k} \coloneqq \{ (T', \bm k') \mid T' \text{ is a subtree of } T \land \bm k' = \bm k \}.
\end{align}
We usually use the deepest perfect $M$-ary tree $T_\mathrm{max}$ (i.e., its depth is $D_\mathrm{max}$) as the representative tree to enlarge the meta-tree. In this paper, we assume $\bm k$ is given and fixed. 

For given $T_\mathrm{max}$, we assume the following prior distribution\cite{suko_alg, MTRF, full_rooted_trees} over the trees in the meta-tree $M_{T_\mathrm{max}, \bm k}$.
\begin{align}
    p(T) = \prod_{s \in \mathcal{I}(T)} g_s \prod_{s' \in \mathcal{L}(T)} (1-g_{s'}), \label{tree_prior}
\end{align}
where $g_s \in [0, 1]$ is a given hyperparameter assigned to each $s \in \mathcal{S}(T_\mathrm{max})$. For any $s \in \mathcal{L}(T_\mathrm{max})$, we assume $g_s = 0$.

Given a tree $T$ in $M_{T_\mathrm{max},\bm k}$, parameters $\bm \theta \coloneqq ( \theta_s )_{s \in \mathcal{L}(T)}$ are independently assigned to each leaf node $s \in \mathcal{L}(T)$ according a prior distribution, i.e., it has the following form.
\begin{align}
    p( \bm \theta | T) = \prod_{s \in \mathcal{L}(T)} p(\theta_s).
\end{align}
Note that $p(\theta_s)$ does not depend on $T$ but only on $s$. In other words, we assume for any $s \in \mathcal{L}(T)$ and $s' \in \mathcal{L}(T')$, $s = s' \Rightarrow p(\theta_s) = p(\theta_{s'})$ holds even if $T \neq T'$.

Finally, given an explanatory variable\footnote{We can assume the explanatory variable is either a given constant or a random variable following any kind of distribution.} $\bm x \in \{ 1, 2, \dots , M\}^K$, an objective variable $y$ is assumed to be generated according to the following distribution.
\begin{align}
    p(y | \bm x, T, \bm \theta) = p(y | \theta_{s_{T, \bm k}(\bm x)}). \label{generative_model}
\end{align}
Here, $s_{T, \bm k}(\bm x)$ is a leaf node of $T$ determined by $\bm x$ in the following procedure. Start from the root node $s_\lambda$, which has a feature assignment index $k_{s_\lambda}$. Check the $k_{s_\lambda}$th element of $\bm x$. If $x_{k_{s_\lambda}} = m$, move to the $m$th child node of $s_\lambda$. Repeat this procedure up to any leaf node. In addition, we assume the marginal likelihood $\int \prod_{i} p(y_i|\theta_s) p(\theta_s) \mathrm{d}\theta_s$ is feasible with an acceptable cost. For example, it is satisfied if $p(y_i|\theta_s)$ is a usual exponential family and $p(\theta_s)$ is its conjugate prior.

In the following sections, we will calculate the posterior distribution $p(T|\bm x^n, \bm y^n)$ given an i.i.d.\ sample $(\bm x^n, y^n) \coloneqq \{ (x_i, y_i) \}_{i=1}^n$ from Eq.\ \eqref{generative_model}. 

\section{Previous studies\cite{suko_alg,MTRF}: sequential updating}

In the previous studies\cite{suko_alg,MTRF}, $p(T | \bm x^n, y^n)$ is sequentially calculated by applying the following proposition for $i = 1, 2, \dots , n$.

\begin{prop}[\cite{suko_alg,MTRF}]\label{prop}
$p(T|\bm x^i, y^i)$ had the following parameterization.
\begin{align}
    p(T|\bm x^i, y^i) = \prod_{s \in \mathcal{I}(T)} g_{s|\bm x^i, y^i} \prod_{s' \in \mathcal{L}(T)} (1-g_{s'|\bm x^i, y^i}).
\end{align}
Here, $g_{s|\bm x^i, y^i} \in [0, 1]$ is updated from $g_{s|\bm x^{i-1}, y^{i-1}}$ of $p(T|\bm x^{i-1}, y^{i-1})$ as follows.
\begin{align}
    g_{s|\bm x^i, y^i} \coloneqq \begin{cases}
        \frac{g_{s|x^{i-1},y^{i-1}}q(y_i|\bm x_i, s_\mathrm{ch})}{q(y_i|\bm x_i, s)}, & s \succ s_{T_\mathrm{max},\bm k}(\bm x_i), \\
        g_{s|\bm x^{i-1}, y^{i-1}}, & \text{otherwise},
    \end{cases}
\end{align}
where $s \succ s_{T_\mathrm{max},\bm k}(\bm x_i)$ means $s$ is an ancestor node of $s_{T_\mathrm{max},\bm k}(\bm x_i)$, $s_\mathrm{ch}$ denotes the child node of $s$ on the path from the root node $s_\lambda$ to the leaf node $s_{T_\mathrm{max}, \bm k}(\bm x_i)$, and $q(y_i|\bm x_i, s)$ is defined for any $s \succeq s_{T_\mathrm{max},\bm k}(\bm x_i)$ as follows.
\begin{align}
    &q(y_i|\bm x_i, s) \coloneqq \\
    &\begin{cases}
        \int p(y_i | \bm x_i, \theta_s) p(\theta_s | \bm x^{i-1}, y^{i-1}) \mathrm{d} \theta_s, & s = s_{T_\mathrm{max},\bm k}(\bm x_i), \\
        (\star), & s \succ s_{T_\mathrm{max},\bm k}(\bm x_i),
    \end{cases} \nonumber \\
    &(\star) = (1 - g_{s|\bm x^{i-1}, y^{i-1}}) \nonumber \\
    &\qquad \times \textstyle \int p(y_i | \bm x_i, \theta_s) p(\theta_s | \bm x^{i-1}, y^{i-1}) \mathrm{d} \theta_s \nonumber \\
    &\qquad \qquad + g_{s|\bm x^{i-1},y^{i-1}} q(y_i | \bm x_i, s_\mathrm{ch}).
\end{align}
\end{prop}

For each $i = 1, 2, \dots , n$, the above formulas are calculated for the nodes on the path from the root node $s_\lambda$ to the leaf node $s_{T_\mathrm{max}, \bm k}(\bm x_i)$. The updating method in Proposition \ref{prop} is suitable to the situation where we have to repeat a data point observation and a prediction without delay. However, the total cost to calculate $p(T | \bm x^n, y^n)$ is $O(n D_\mathrm{max})$ because the length of each path is $D_\mathrm{max}$.

\section{Main results: batch updating}

First, we define $\bm x_s$ as $\{ \bm x_i \}_{i: s \succeq s_{T_\mathrm{max}, \bm k}(\bm x_i)}$. Namely, $\bm x_s$ is the set of data points that pass through $s$ in the data generating process, and $\bigcup_{s \in \mathcal{L}(T)} \bm x_s = \bm x^n$ holds for any $T$ in the meta-tree $M_{T_\mathrm{max}, \bm k}$. In a similar manner, we define $y_s \coloneqq \{ y_i \}_{i: s \succeq s_{T_\mathrm{max}, \bm k}(\bm x_i)}$. By using these notations, the batch updating formula for $g_{s | \bm x^n, y^n}$ is represented as follows.

\begin{theo}\label{main_result}
For any $s \in \mathcal{S}(T_\mathrm{max})$, the following holds.
\begin{align}
    g_{s|\bm x^n, y^n} = 
    \begin{cases}
        \frac{g_s \prod_{s' \in \mathrm{Ch}(s)}q(y_{s'}|\bm x_{s'}, s')}{q(y_s|\bm x_s, s)}, & s \in \mathcal{I}(T_\mathrm{max}), \\
        g_s, & s \in \mathcal{L}(T_\mathrm{max}),
    \end{cases} \label{update_g}
\end{align}
where $\mathrm{Ch}(s)$ denotes the set of child nodes of $s$ on $T_\mathrm{max}$ and $q(y_s|\bm x_s, s)$ is defined for any $s \in \mathcal{S}(T_\mathrm{max})$ as follows.
\begin{align}
    &q(y_s|\bm x_s, s) \nonumber \\
    &\coloneqq \begin{cases}
        \int p(y_s | \bm x_s, \theta_s) p(\theta_s) \mathrm{d} \theta_s, & s \in \mathcal{L}(T_\mathrm{max}), \\
        (*), & s \in \mathcal{I}(T_\mathrm{max}),
    \end{cases} \label{qs} \\
    &(*) = (1-g_s) \textstyle\int p(y_s | \bm x_s, \theta_s) p(\theta_s) \mathrm{d} \theta_s\nonumber \\
    &\qquad \quad + g_s \textstyle \prod_{s' \in \mathrm{Ch}(s)} q(y_{s'} | \bm x_{s'}, s'),
\end{align}
where $p(y_s | \bm x_s, \theta_s) = \prod_{i: s \succeq s_{T_\mathrm{max}, \bm k}(\bm x_i)} p(y_i | \bm x_i, \theta_s)$. Note that, $\bm x_s$ and $y_s$ may be empty. In such a case, we define $p(y_s | \bm x_s, \theta_s) = 1$, which is similar to usual empty products.
\end{theo}

\noindent \textit{Proof:} In \cite{full_rooted_trees}, properties of the tree prior distribution \eqref{tree_prior} is summarised. In particular, a sufficient condition to efficiently calculate the posterior distribution $p(T|z)$ given some kind of data $z$ observed according to a distribution $p(z|T)$ that determined by the tree $T$. The condition, which is known as Condition 3 in \cite{full_rooted_trees}, is as follows.
\begin{align}
    p(z|T) = \prod_{s \in \mathcal{I}(T)} f(z,s) \prod_{s' \in \mathcal{L}(T)} h(z,s') \label{Condition3}
\end{align}

Although our model has an additional parameter $\bm \theta$, we can marginalize it and the model is represented as follows.
\begin{align}
    p(y^n | \bm x^n, T) = \prod_{s \in \mathcal{L}(T)} \int p(y_s | \bm x_s, \theta_s) p(\theta_s) \mathrm{d}\theta_s. \label{marginal_likelihood}
\end{align}
This has an equivalent form to \eqref{Condition3} because we can represent it by substituting $z = y^n$, $f(y^n,s) = 1$, and $h(y^n,s) = \int p(y_s | \bm x_s, \theta_s) p(\theta_s) \mathrm{d}\theta_s$. Therefore, Theorem 7 in \cite{full_rooted_trees} can be applied to our model. Then, Theorem \ref{main_result} in this paper straightforwardly holds. Here, we describe only an overview of the proof and the meaning of Theorem 7 in \cite{full_rooted_trees}.

From the Bayes' theorem and Eqs.\ \eqref{tree_prior} and \eqref{marginal_likelihood}, we have the following.
\begin{align}
    &p(T | \bm x^n, y^n) =\frac{1}{p(y^n | \bm x^n)} \prod_{s \in \mathcal{I}(T)} g_s \nonumber \\
    &\quad \times \prod_{s' \in \mathcal{L}(T)} (1-g_{s'}) \int \! p(y_{s'} | \bm x_{s'}, \theta_{s'}) p(\theta_{s'}) \mathrm{d}\theta_{s'}
\end{align}
Therefore, the posterior distribution has the same form as the prior distribution \eqref{tree_prior} except the normalization term $p(y^n | \bm x^n)$, which is the likelihood marginalized over the metatree $M_{T_\mathrm{max},\bm k}$. Theorem 7 in \cite{full_rooted_trees} provides the way to calculate the normalization term and distribute it to each factor to re-parameterize the posterior distribution. Eq.\ \eqref{qs} corresponds to the algorithm to calculate the normalization term (actually, $p(y^n | \bm x^n) = q(y_{s_\lambda}|\bm x_{s_\lambda}, s_\lambda)$ holds). Eq.\ \eqref{update_g} corresponds the re-parameterization.
\hfill $\Box$

The updating formulas in Theorem \ref{main_result} are applied for all $s \in \mathcal{S}(T_\mathrm{max})$. Then, the computational cost is $O(|\mathcal{S}(T_\mathrm{max})|)$. 
Since $|\mathcal{S}(T_\mathrm{max})|$ is independent of $n$, this will be more efficient than the previous method when $n$ is large. Moreover, we can reuse $\int p(y_s | \bm x_s, \theta_s) p(\theta_s) \mathrm{d} \theta_s$ for another meta-tree $M_{T_\mathrm{max}, \bm k'}$ when the feature indices assigned to the ancestor nodes of $s$ in $M_{T_\mathrm{max}, \bm k'}$ are the same as those in $M_{T_\mathrm{max}, \bm k}$.

However, if $n$ is small, e.g., each data point is observed one by one, this method is not efficient in the above form. Therefore, we further improve this method.

If $\bm x_s = \emptyset$ holds, then $q(y_{s'}|\bm x_{s'}, s') = 1$ holds for any descendant node $s'$ of $s$. Therefore, the following holds.
\begin{coro}\label{coro}
Eqs.\ \eqref{update_g} and \eqref{qs} can be represented as follows.
\begin{align}
    &g_{s|\bm x^n, y^n} = \begin{cases}
        g_s, &  \bm x_s = \emptyset,\\
        \frac{g_s \prod_{s' \in \mathrm{Ch}(s)}q(y_{s'}|\bm x_{s'}, s')}{q(y_s|\bm x_s, s)}, & \text{otherwise},
    \end{cases}\\
    &q(y_s|\bm x_s, s) = \begin{cases}
        \int p(y_s | \bm x_s, \theta_s) p(\theta_s) \mathrm{d} \theta_s, & \bm x_s = \emptyset, \\
        (*), & \text{otherwise}.
    \end{cases}
\end{align}
\end{coro}

Let $\mathcal{S}_{\bm x^n}$ denote the set of all nodes where $\bm x_s \neq \emptyset$ holds. These functions are called $|\mathcal{S}_{\bm x^n}|$ times. Moreover, $|\mathcal{S}_{\bm x^n}| \leq n D_\mathrm{max}$ holds. When $n=1$, the equality holds and the updating formulas coincide with those in Proposition \ref{prop}. Therefore, the method in Corollary \ref{coro} is always more efficient than that in Proposition \ref{prop} and its complexity is bounded above by $O(|\mathcal{S}(T_\mathrm{max})|)$.

Further, if we impose a stronger assumption that $p(\theta_s) = p(\theta_{s'})$ for any nodes $s$ and $s'$, and once $\bm x_s$ is concentrated at one point, i.e., 
\begin{align}
(\bm x_i = \bm x_j \text{ for any } \bm x_i, \bm x_j \in \bm x_s) \text{ or } (\bm x_s = \emptyset), \label{concentrate}
\end{align}
then either $q(y_{s'}|\bm x_{s'}, s') = q(y_s|\bm x_s, s)$ or $q(y_{s'}|\bm x_{s'}, s') = 1$ holds for any descendant node $s'$ of $s$. Therefore, the following holds.
\begin{coro}\label{coro2}
If $p(\theta_s) = p(\theta_{s'})$ holds for any nodes $s$ and $s'$, Eqs.\ \eqref{update_g} and \eqref{qs} can be represented as follows.
\begin{align}
    &g_{s|\bm x^n, y^n} = \begin{cases}
        g_s, &  \eqref{concentrate},\\
        \frac{g_s \prod_{s' \in \mathrm{Ch}(s)}q(y_{s'}|\bm x_{s'}, s')}{q(y_s|\bm x_s, s)}, & \text{otherwise},
    \end{cases}\\
    &q(y_s|\bm x_s, s) = \begin{cases}
        \int p(y_s | \bm x_s, \theta_s) p(\theta_s) \mathrm{d} \theta_s, & \eqref{concentrate}, \\
        (*), & \text{otherwise}.
    \end{cases}
\end{align}
\end{coro}
These formulas mean that we need not fix the maximum depth of trees any more. Instead, we can call the above recursive functions until $\bm x_s$ is concentrated at one point.

\section{Experiments}

Table \ref{table} shows CPU times (avg.\ $\pm$ std.\ in 100 tests) of the method in \cite{suko_alg,MTRF} and the proposed method (available in \cite{bayesml}).

\begin{table}[htbp]
    \centering
    \caption{CPU time (msec) comparison where $M=2$, $K=5$, $D_\mathrm{max}=5$, $p(y|\theta_s)$ is the Bernoulli distribution, and $p(\theta_s)$ is the beta distribution.}
    \label{table}
    \begin{tabular}{c|ccc}
        sample size $n$ & 50 & 100 & 200 \\
        \hline
        Previous\cite{suko_alg,MTRF} & $1.67 \pm 0.20$ & $3.29 \pm 0.27$ & $6.72 \pm 0.85$\\
        Proposal\cite{bayesml} & $1.78 \pm 0.41$ & $1.89 \pm 0.23$ & $1.90 \pm 0.22$\\
        \hline
    \end{tabular}
\end{table}



\end{document}